\documentclass{article}

\usepackage[preprint]{nips_2018}
\usepackage{wrapfig}

\usepackage{tikz,pgfplots}

\usepackage{tcolorbox}

\usepackage[utf8]{inputenc} 
\usepackage[T1]{fontenc}    
\usepackage{hyperref}       
\usepackage{url}            
\usepackage{booktabs}       
\usepackage{nicefrac}       
\usepackage{microtype}      
\usepackage{graphicx}
\usepackage{times}
\usepackage[english]{babel}
\usepackage{bbm}
\usepackage{amsmath}
\usepackage{amsfonts}
\usepackage{ntheorem}

\usepackage{lscape} 
\usepackage{geometry}
\usepackage{breqn}
\usepackage[]{algorithm2e}
\usepackage{algpseudocode}
\usepackage{subfigure}
\usepackage{epsfig}
\usepackage{overpic}
\usepackage{rotating}
\usepackage{here}
\usepackage{txfonts}
\usepackage{color}
\usepackage[mathscr]{eucal}
\usepackage{tikz}
\usepackage{natbib}

\newcommand{\X}{\mathcal{X}}
\newcommand{\Y}{\mathcal{Y}}

\DeclareMathAlphabet{\mathpzc}{OT1}{pzc}{m}{it}
\DeclareFontFamily{OT1}{pzc}{}
\DeclareFontShape{OT1}{pzc}{m}{it}{<-> s * [0.900] pzcmi7t}{}
\DeclareMathAlphabet{\mathpzc}{OT1}{pzc}{m}{it}

\newcommand{\Inp}{\mathcal{X}}
\newcommand{\Outp}{\mathcal{Y}}
\newcommand{\obs}{\mathbf{x}}
\newcommand{\z}{\mathbf{z}}
\newcommand{\y}{\mathbf{y}}

\newcommand{\bit}{\begin{itemize}}
\newcommand{\eit}{\end{itemize}}

\newcommand{\argmin}{\mathop{\mathrm{arg\,min}}}

\newcommand{\Dist}{\mathcal{D}}

\def\Prodi{\mathop{{\lower 3pt\hbox{\epsfxsize=15pt\epsfbox{pi.ps}}}}}
\def\prodi{\mathop{{\lower 1pt\hbox{\epsfxsize=8pt\epsfbox{pi.ps}}}}}

\newcommand{\beq}{\begin{equation}}
\newcommand{\eeq}{\end{equation}}

\newcommand{\bea}{\begin{eqnarray}}
\newcommand{\eea}{\end{eqnarray}}
\newcommand{\beas}{\begin{eqnarray*}}
\newcommand{\eeas}{\end{eqnarray*}}

\newcommand{\Trn}{\mathcal{S}}
\newcommand{\CC}{\texttt{CC}}

\newcommand{\BBe}{\mathbb{E}}
\newcommand{\BBp}{\mathbb{P}}

\newtheorem{theo}{Theorem}

\newtheorem {pro}{Proposition}
\newtheorem{defi}{Definition}

\newdimen\AAdi%
\newbox\AAbo%
\def\AAk#1#2{\setbox\AAbo=\hbox{#2}\AAdi=\wd\AAbo\kern#1\AAdi{}}%
\def\AAr#1#2#3{\setbox\AAbo=\hbox{#2}\AAdi=\ht\AAbo\raise#1\AAdi\hbox{#3}}%
\title{Rademacher Generalization Bounds for \\ Classifier Chains}

\author{
  Moura Simon\\
  University Grenoble Alps\\
  LIG - CNRS \\
  \small{\texttt{simon.moura|at|imag.fr}}\\
  \And
  Massih-Reza Amini\\
  University Grenoble Alps\\
  LIG - CNRS \\
  \small{\texttt{massih-reza.amini|at|imag.fr}}\\
  \And
  Sana Louhichi\\
  University Grenoble Alps\\
  LJK - CNRS \\
  \small{\texttt{sana.Louhichi|at|imag.fr}}\\
  \And
  Marianne Clausel \\
  Institut Elie Cartan de Lorraine\\
  \small{\texttt{marianne.clausel|at|univ-lorraine.fr}}\\
}

\begin{document}

\maketitle

\begin{abstract}
In this paper, we propose a new framework to study the generalization property of classifier chains trained over observations associated with multiple and interdependent class labels. The results are based on large deviation inequalities for Lipschitz functions of weakly dependent sequences proposed by \cite{RIO2000905}. We believe that the resulting generalization error bound brings many advantages and could be adapted to other frameworks that consider interdependent outputs. First, it explicitly exhibits the dependencies between class labels. Secondly, it provides insights of the effect of the order of the chain on the algorithm generalization performances. Finally, the two dependency coefficients that appear in the bound could also be used to design new strategies to decide the order of the chain.
\end{abstract}

\section{Introduction}

Mutli-label classification consists in associating  observations, described in some input space $\Inp\subseteq\mathbb{R}^d$ to a subset of class labels $\Outp_\obs\subseteq\Outp, \forall \obs\in\Inp$. The framework  has received increased interest from industry and academic research communities during the past years, as it encompasses many domain applications, such text classification \cite{Shantanu:04, Rubin:2012} or medical diagnosis \cite{Vens2008, Zhang:2006}. Various classification techniques have been proposed for the learning of efficient prediction models in the multi-label learning case. The baseline approach, referred as Binary Relevance (\texttt{BR}) \cite{Luaces2012, Read2014}, consists in transforming the original problem into $|\Outp|$ binary problems and to train one independent classifier per problem. The inherent assumption in \texttt{BR} is that class labels are independent which may not always hold in practice. However, different studies have provided empirical evidence that taking into account the dependencies between classes may lead to a substantial increase of the overall test performance \cite{Montanes:2014, Rauber:14, Li:2014}. The classifier chain algorithm is a popular approach that take advantage of the class dependencies \cite{Read:2011}. It consists in learning $|\Outp|$ binary classifiers sequentially over the  expanded feature space, formed by the concatenation of the initial input space with the previous classifiers predictions. Although the dependencies introduced between the original input space and the expanded space is not defined explicitly, many studies have shown the efficiency of \texttt{CC} over \texttt{BR} even with a random choice of the classifiers considered in the chain \cite{Read:2011}.  Up to our knowledge all these works have not yet considered  the consistency of the Empirical Risk Minimization (\texttt{ERM}) principle \cite{Vapnik:1991} with interdependent observations for the training of a new classifier at each step of the algorithm.

\textbf{Contributions.} In this paper, we propose a theoretical study that tackles this problem. Our approach is based on large deviation inequalities of Rio \cite{RIO2000905}, that extend McDiarmid inequalities  \cite{McDiarmid:89} to the case of dependently distributed observations, characterized by Lipschitz functions of some weakly dependent sequences. We provide generalization error bounds for each classifier in the chain trained on observations where the feature vector representations are expanded with the previous classifier predictions. The sequence of classes may be deduced by a Markov chain of order one which links the class labels by class transition probabilities estimated over a training set.
The resulting bounds involve a Rademacher type of complexity term defined over training samples where examples are not independently distributed, as well as dependency coefficients that measure the variational distance between the probability measure of the current class conditionally to the whole training examples and the previous class labels in the chain.

\textbf{Organization.}
The remainder of the paper is organized as follows: Section \ref{Notations} introduces the framework used throughout this paper. In Section \ref{Bounds}, we present the Rademacher complexity bounds for \texttt{CC}. Finally, we conclude this study in Section \ref{Conclusion}.

\section{General Framework and Definitions}
\label{Notations}

In this section we define the {\CC} framework that will be investigated in this paper. Throughout this paper, we suppose that for any positive integer $n$, $[n]$ denotes the set $[n]=~\{1, \ldots , n\}$, and for any vector $\mathbf{q}=(q_1,\cdots,q_n)\in \mathbb{R}^n$ and any $s$-uplet $J=(j_1,\cdots,j_s)\in [n]^s$; $q_J$ denotes $(q_{j_1},\cdots,q_{j_s})$. Let the output space be defined as $\Y=\{-1,+1\}^K$, where $K$ is the number of class labels, and suppose that each random pair of example $(X,\boldsymbol{Y}=(Y^{(k)})_{k\in [K]})\in\X\times\Y$ is identically and independently distributed  with respect to a fixed yet unknown probability distribution $\Dist$, where $Y^{(k)}\in\{-1,+1\}$ is the $k^{th}$ class membership indicator.  Furthermore, we consider a labeled sample $\Trn=(\obs_i,\y_i)_{1\le i\le m}\in (\X\times\Y)^m$ of size $m$ drawn i.i.d. according to $\Dist$, and $K$ $\{-1,+1\}$-valued class of functions $(\mathcal{F}_k)_{1\le k\le K}$, linked along a fixed chain where each class $\mathcal{F}_k=\{f_k: \bar{\X}_k\rightarrow \{-1,+1\}\}$ is defined over  the input space augmented by the predictions of the $k-1$ previous classifiers in the chain; $\bar{\X}_k=\X\times \{-1,+1\}^{k-1}$. By convention, we set $\bar{\X}_1=\X$.

\subsection{{\CC} algorithm and Learning Objective}

The objective consists in  learning sequentially $K$ binary classifiers over the augmented input spaces, in a way that at each iteration $k\in [K]$, the association between observations represented in $\bar{\X}_k$ and the $k^{th}$ class in the sequence of labels is found with a low generalization error defined as
\begin{equation*}
{\cal L}(f_k)= \mathbb{E}_{(Z^{(k)},y^{(k)})\sim \bar\Dist_k}[\mathbbm{1}_{f_k(Z^{(k)})\neq y^{(k)}}], \forall f_k\in\mathcal F_k,
\end{equation*}

where $\bar\Dist_k$ is a probability distribution over $\bar{\X}_k\times \{-1,+1\}$,  $\mathbbm{1}_\pi$ \begin{wrapfigure}[12]{r}{0.46\textwidth}

\vspace{-2mm}\tcbset{width=0.46\columnwidth,before=,after=, colframe=black,colback=white, fonttitle=\bfseries, coltitle=black, colbacktitle=white, boxrule=0.2mm,arc=1mm, left = 2pt}

\begin{tcolorbox}[title={\CC} Algorithm, label={algo:CC}]
\textbf{Input:} A training set $\Trn=(\obs_i,\y_i)_{i\in [m]}$\\
\textbf{Initialization:} Set $\z^{(1)}_i=\obs_i, \forall i\in[m]$\\
\For{$k=1,\ldots,K$}{

     \textbf{Train} $\hat{f}_k=\argmin_{f_k\in\mathcal{F}_k} \mathcal{\hat{L}}_m(f_k)$;\\
     Set $\z^{(k+1)}_i=(\z^{(k)}_i, \hat{f}_k(\z_i^{(k)})); \forall i\in[m]$
}

\textbf{Test:} For a new example $\z^{(1)}\!=\!\obs;$ ~~~~~~~~~~~\begin{center}predict $\left(\hat f_k(\z^{(k)})\right)_{k\in[K]}$
\end{center}
\end{tcolorbox}

\end{wrapfigure}the indicator function that is equal to $1$ if the predicate $\pi$ is true and $0$ otherwise, and  for any random pair of example $(X,\boldsymbol{Y}=(Y^{(k)})_{k\in [K]})\in\X\times\Y$; $(Z^{(k)},Y^{(k)})\in \bar{\X}_k\times \{-1,+1\}$ is the associated pair, with $Z^{(k)}=(X,f_1(X),\ldots,f_{k-1}(Z^{(k-1)}))$ the vector formed by $X$ augmented by predictions provided by the $k-1$ previous classifiers.

Following the Empirical Risk Minimization (\texttt{ERM}) principle \cite{Vapnik:1991}, the learning of the chain classifiers is carried out sequentially. At each iteration $k\in [K]$ the learner receives a labeled sample $\Trn_k=(\z^{(k)}_i,y^{(k)}_i)_{1\le i\le m}\in (\bar{\X}_k\rightarrow \{-1,+1\})^m$, based on the initial training set $\Trn$ where $\forall i\in [m], \z^{(1)}_i=\obs_i$; and find the target labeling function $f_k\in \mathcal{F}_k$ by minimizing the empirical error defined by
\begin{equation*}
\mathcal{\hat{L}}_m(f_k)=\frac{1}{m}\sum_{i=1}^m \mathbbm{1}_{ f_k(\z_i^{(k)})\neq y_i^{(k)}}.
\end{equation*}

At the next iteration, the feature representation of observations is augmented by the predictions of the current classifier: $\forall i, \z_i^{(k+1)}=\left(\z_i^{(k)}, \hat f_k(\z_i^{(k)})\right)$. For a new example $\obs\in\Inp$, the prediction is made using the learned sequence of predictors $\left(\hat f_k(\z^{(k)}\right)_{k\in[K]}$.

\subsection{Dependency measures}
Note that after the first iteration, $k\ge 2$, the observations in $\Trn_k$ are no longer independently distributed with respect to $\bar\Dist_k$. Indeed, the predictors $\hat f_k$ are found using all the training samples making the new representation of observations to be interdependent. In other words, for each example $(\obs_i,\y_i)\in\Trn$, the corresponding instance $\z_i^{(k)}$ is deduced from the vector $\left(\obs_i,\left(\obs_j,y_j^{(1)},\ldots,y_j^{(k-1)}\right)_{j\in [m]}\right)$.   \\

In this work, we study the generalization ability of the above \texttt{ERM} principle that outputs a predictor $\hat{f}_k\in \mathcal{F}_k$ at each iteration $k\in [K]$ of the algorithm by considering two measures of dependency based on the total variation distance between the following probability measures \cite{Dudley2010}.

\begin{defi}
\label{defalphabeta}
For any $k\in\{2,\ldots,K\}$ and for any  $\ell\in [m]$ ; we define the following global and local dependency measures 
\begin{equation}
\label{eq:gamma}
\rho^{(k)}=\sup_{(\obs_{[m]},\left(y_{[m]}^{(s)}\right)_{s\in[k-1]})}\sum_{y_i^{(k)}}\left|\BBp\left(y_{i}^{(k)}\left|\right.\left(\obs_{[m]},\left(y_{[m]}^{(s)}\right)_{s\in [k-1]}\right)\right)-\BBp\left(y_{i}^{(k)}\right)\right|,
\end{equation}
and
\begin{equation}
\label{eq:delta}
\gamma_{\ell}^{(k)}=
\sup_{(\obs_{I_\ell},(y_{I_\ell}^{(s)})_{s\in[k-1]})}\sum_{y_{I_\ell}^{(k)}}\left|\BBp\left(y_{I_\ell}^{(k)}\left|\right.\left(\obs_{I_\ell},\left(y_{I_\ell}^{(s)}\right)_{s\in [k-1]}\right)\right)-\BBp\left(y_{I_\ell}^{(k)}\right)\right|,
\end{equation}
where, $I_\ell=\{\ell+1,\ldots,m\}\subseteq [m]$ is a set of $m-\ell+1$ indexes. 
\end{defi}
Hence at each iteration $k\in\{2,\ldots,K\}$ of the {\CC} algorithm, the coefficient $\rho^{(k)}$ measures the variational distance between the probability measure of the $k^{th}$ class conditionally to the whole training examples and their previous class labels in the chain and the distribution of the $k^{th}$ class; while the coefficient $\gamma_\ell^{(k)}$ is the variational distance between the same probability measures but over any subset of the training samples of size   $m-\ell+1$. Note that if at each step, the $k$-th label $(y_i^{(k)})$ is independent from the whole set of augmented inputs $(\z^{(k)}_i)$ at step $k$ these measures are all equal to zero.

\noindent We now consider an important framework for \CC{}s and give explicit expressions of these dependence measures in this case.\\

\textbf{The Markovian setting.} Here, we consider the case where the chaining in the \CC{} is defined over a Markov Chain . Let us assume that for any subsets $I\subseteq I' \subseteq [m]$
\begin{eqnarray}\label{A}
&&\BBp\left(y_{I}^{(k)}\left|\right. \left(\obs_{I'},\left(y_{I^{'}}^{(s)}\right)_{1\leq s\leq k-1}\right)\right)=\BBp\left(y_{I}^{(k)}\left|\right.y_{I}^{(k-1)}\right).
\end{eqnarray}
Then for any $k\geq 2$, we get 
\[
\rho^{(k)}=\gamma_1^{(k)},
\] 
and for any $\ell\in[m]$ and $I_\ell=\{\ell+1,\ldots,m\}$
\begin{eqnarray*}
\gamma_{\ell}^{(k)}=\sup_{I_\ell}\sum_{(y_{I_\ell}^{(k)})}
\left|\prod_{j\in {I_\ell}}\BBp\left(y_j^{(k)}\left|\right.y_j^{(k-1)}\right)\right.\left.-\prod_{j\in {I_\ell}}\BBp\left(y_j^{(k)}\right)\right|.
\end{eqnarray*}
Note, that replacing all the probabilities above with empirical one, allows to propose an estimation procedure for these coefficients from the data.

From this definition we have the following proposition which controls the deviation of $\sup_{f_k \in\mathcal F_k} \left({{\cal L}}(f_k) - \hat{\cal L}_m(f_k)\right)$.

\begin{pro}\label{lem1}
Let $(X,\boldsymbol{Y}=(Y^{(k)})_{k\in[K]})\in\Inp\times\Outp$ random pairs of examples drawn i.i.d with respect to a probability distribution $\Dist$. For any $k\in\{2,\ldots,K\}$, let $\bar\Dist_k$ be the probability distribution of the vector $(Z^{(k)},Y^{(k)})\in \bar{\X}_k\times \{-1,+1\}$; with $Z^{(k)}$ the vector formed by $X$ augmented by predictions provided by the $k-1$ previous classifiers of the \CC{} algorithm. Let $\Trn_k=(\z^{(k)}_i,y^{(k)}_i)_{1\le i\le m}\in (\bar{\X}_k\rightarrow \{-1,+1\})^m$ be a dataset of $m$ examples drawn from $\bar\Dist_k$ and obtained by extending the feature representations of an original training set $\Trn$, then for any $\epsilon>0$ we have~:
\begin{equation*}
\BBp\left(\sup_{f_k \in\mathcal F_k} \left({{\cal L}}(f_k)-\hat{\cal L}_m(f_k)\right)-\BBe_{\Trn_k}\left[\sup_{f_k \in \mathcal F_k} \left({{\cal L}}(f_k)\right)-\hat{\cal L}_m(f_k) \right]\geq \epsilon\right) \leq \exp\left(- \frac{2m^2{\epsilon}^2}{\sum_{\ell=1}^{m}(1+ 2m\gamma_{\ell+1}^{(k)})^2}\right).
\end{equation*}
\end{pro}

The proof is based on the following concentration inequality of  \cite{RIO2000905} which states that~:
\begin{theo}[\cite{RIO2000905}]
\label{rio}
Let $T_1,\ldots, T_m$ be random variables (not necessarily \textit{independent}) with values in some metric spaces $\otimes_{i=1}^m E_i$, and assume that $\Phi$ is a real-valued function defined on $E_1\times E_2\times\cdots\times E_m$. furthermore, for $\ell\in \{1,\cdots,m\}$, let $f_\ell$ be a finite real-valued function defined on $E_1\times\cdots\times E_\ell$, by
 \[
 f_\ell (t_1,\cdots,t_\ell)=  \BBe\left[\Phi(T_1,\cdots,T_m)| T_1=t_1,\cdots,T_\ell=t_\ell\right],
 \]
 for any $(t_1,\cdots,t_\ell)\in E_1\times\cdots\times E_\ell$.
 Suppose that, for each $\ell\in \{1,\cdots,m\}$, there exist two positive constants $\eta_\ell$ and $\xi_\ell$ and two real-valued functions $h_\ell$ and $g_\ell$ defined  on $E_1\times\cdots\times E_{\ell-1}$ such that, almost surely,
 \begin{equation}\label{eq1}
g_\ell(T_1,\cdots,T_{\ell-1})-\eta_\ell  \leq f_\ell(T_1,\cdots,T_\ell) \leq h_\ell(T_1,\cdots,T_{\ell-1})+ \eta_\ell
 \end{equation}
 with
  \begin{equation}\label{eq2}
  h_\ell(T_1,\cdots,T_{\ell-1})-g_\ell(T_1,\cdots,T_{\ell-1})\leq \xi_\ell.
 \end{equation}
Then for all $\epsilon>0$
 \[
 \BBp\left(\Phi(T_1,\cdots,T_m)-\BBe\left[\Phi(T_1,\cdots,T_m)\right]\geq \epsilon\right) \leq \exp\left(- \frac{2\epsilon^2}{\sum_{\ell=1}^m(\xi_\ell+ 2\eta_\ell)^2}\right).
 \]
 \end{theo}

\textbf{Proof of Proposition \ref{lem1}}.
Let $\Phi:\mathbb{R}^m\rightarrow \mathbb{R}$ be the following multivariate function 
\begin{equation*}
\Phi(\mathbf{t}_1^{(k)},\cdots,\mathbf{t}_m^{(k)})=\sup_{f_k \in\mathcal{F}_k} \left({{\cal L}}(f_k)-\hat{\cal L}_m(f_k)\right)=\sup_{f_k \in\mathcal{F}_k}\frac{1}{m}\sum_{i=1}^m \left(\BBe[T_i^{(k)}]-\mathbf{t}_i^{(k)}\right),
\end{equation*}
where for any $f_k\in\mathcal{F}_k$; $T^{(k)}=\mathbbm{1}_{f_k(Z^{(k)})\neq y^{(k)}}$ is the 0/1 loss of the random variable $(Z^{(k)},y^{(k)})$ generated from the distribution $\bar \Dist_k$ and $\forall i\in[m], \mathbf{t}_i^{(k)}=\mathbbm{1}_{f_k(\z_i^{(k)})\neq y_i^{(k)}}$ is the 0/1 loss of a training example $(\z^{(k)}_i,y_i^{(k)})\in\Trn_k$.   As $\left(X_i,Y_i^{(1)},\cdots,Y_i^{(k)}\right)_{i\in[\ell]}$ is independent of the couple of vectors $\left(\left(Y_i^{(k)}\right)_{\ell+1\leq i\leq m},
\left(X_i,Y_i^{(1)},\cdots,Y_i^{(k-1)}\right)_{\ell+1\leq i\leq m}\right)$, and  since $(T_1^{(k)},\cdots,T_\ell^{(k)})$ belongs to the sigma-algebra formed by the set $((X_j,Y_j^{(1)},\cdots,Y_j^{(k)})_{j\in [\ell]}, (X_j,Y_j^{(1)},\cdots,Y_j^{(k-1)})_{\ell+1\leq j\leq m})$ we have
\begin{equation*}\label{in}
\BBe_{(Y_j^{(k)})_{\ell+1\leq j\leq m}}\left[\Phi(T_1^{(k)},\cdots,T_m^{(k)})\mid T_1^{(k)}=\mathbf{t}^{(k)}_1,\cdots,T_l^{(k)}=\mathbf{t}^{(k)}_\ell\right] = \!\!\!\!{\sum_{(y_j^{(k)})_{\ell+1\leq j\leq m}}\!\!\!\! \Phi(\mathbf{t}_1^{(k)},\cdots,\mathbf{t}_m^{(k)}) \BBp\left((y_j^{(k)})_{\ell+1\leq j\leq m}|{\cal E} \right), {\nonumber}}
\end{equation*}
where ${\cal E}$ is the event $\left(\obs_j,(y_j^{(s)})_{s\in [k-1]}\right)_{l+1\leq j\leq m}$. Similarly,
\begin{equation*}\label{e2}
\BBe_{(Y_j^{(k)})_{\ell+1\leq j\leq m}}\left[\Phi(\mathbf{t}_1^{(k)},\ldots,\mathbf{t}_\ell^{(k)}, T_{\ell+1}^{(k)},\cdots, T_m^{(k)})\right]
= \!\!\sum_{(y_j^{(k)})_{l+1\leq j\leq m}}\!\! \Phi(\mathbf{t}_1^{(k)},\cdots,\mathbf{t}_m^{(k)}) \BBp((y_j^{(k)})_{\ell+1\leq j\leq m}).
\end{equation*}

From the equations above it comes
\begin{align}\label{e4}
&\BBe\left[\Phi(T_1^{(k)},\cdots,T_m^{(k)})\mid T_1^{(k)}=\mathbf{t}^{(k)}_1,\cdots,T_l^{(k)}=\mathbf{t}^{(k)}_\ell\right]- \BBe\left[\Phi(\mathbf{t}_1^{(k)},\ldots,\mathbf{t}_\ell^{(k)}, T_{\ell+1}^{(k)},\cdots, T_m^{(k)})\right]\nonumber\\
&= \sum_{(y_j^{(k)})_{l+1\leq j\leq m}} \Phi(\mathbf{t}_1^{(k)},\cdots,\mathbf{t}_m^{(k)})\Delta\left((y_j^{(k)})_{\ell+1\leq j\leq m}\right),
\end{align}
where
\begin{equation*}
\Delta((y_j^{(k)})_{l+1\leq j\leq m})= \BBp((y_j^{(k)})_{l+1\leq j\leq m}|(\obs_j,(y_j^{(s)})_{s\in [k-1]})_{\ell+1\leq j\leq m})-\BBp((y_j^{(k)})_{\ell+1\leq j\leq m}).
\end{equation*}
Consequently, using equations \eqref{eq:delta}, \eqref{e4}; the triangle inequality and  the fact that the function $\Phi$ is always bounded by $1$, we get
\begin{equation}\label{e6}
\left|\BBe\left[\Phi(T_1^{(k)},\cdots,T_m^{(k)})\mid T_1^{(k)}=\mathbf{t}^{(k)}_1,\cdots,T_l^{(k)}=\mathbf{t}^{(k)}_\ell\right]- \BBe\left[\Phi(\mathbf{t}_1^{(k)},\ldots,\mathbf{t}_\ell^{(k)}, T_{\ell+1}^{(k)},\cdots, T_m^{(k)})\right]\right|\leq \gamma^{(k)}_{\ell+1}.
\end{equation}
The above inequality ensures the validity of conditions (\ref{eq1}) and (\ref{eq2}) of Theorem \ref{rio} \cite{RIO2000905}. Indeed, let 
\begin{equation*}
f_\ell(\mathbf{t}_1,\cdots,\mathbf{t}_\ell)= \BBe\left[\Phi(T_1^{(k)},\cdots,T_m^{(k)})| T_1^{(k)}=\mathbf{t}^{(k)}_1,\cdots,T_\ell^{(k)}=\mathbf{t}^{(k)}_\ell\right], \forall \ell\in[m].
\end{equation*}

We have from equation (\ref{e6}),
\begin{equation*}
\BBe\left[\Phi(\mathbf{t}_1^{(k)},\ldots,\mathbf{t}_\ell^{(k)}, T_{\ell+1}^{(k)},\cdots, T_m^{(k)})\right]-\gamma_{\ell+1}^{(k)}\leq  f_\ell(\mathbf{t}_1,\cdots,\mathbf{t}_\ell) \leq \BBe\left[\Phi(\mathbf{t}_1^{(k)},\ldots,\mathbf{t}_\ell^{(k)}, T_{\ell+1}^{(k)},\cdots, T_m^{(k)})\right]+\gamma_{\ell+1}^{(k)}.
\end{equation*}
which satisfies condition (\ref{eq1}), and since
\begin{equation*}
g_\ell(\mathbf{t}_1,\cdots,\mathbf{t}_{\ell-1})-\gamma_{\ell+1}^{(k)}\leq  f_\ell(\mathbf{t}_1,\cdots,\mathbf{t}_\ell)  \leq h_\ell(\mathbf{t}_1,\cdots,\mathbf{t}_{\ell-1})+\gamma_{\ell+1}^{(k)},
\end{equation*}
where
\begin{eqnarray*}
&& h_\ell(\mathbf{t}_1,\cdots,\mathbf{t}_{\ell-1})= \sup_{z\in \{0,1\}}\BBe\left[\Phi\left(\mathbf{t}_1,\cdots,\mathbf{t}_{\ell-1},z, T_{\ell+1}^{(k)},\cdots, T_m^{(k)}\right)\right]\\
&& g_\ell(\mathbf{t}_1,\cdots,\mathbf{t}_{\ell-1})= \inf_{z\in \{0,1\}}\BBe\left[\Phi\left(\mathbf{t}_1,\cdots,\mathbf{t}_{\ell-1},z, T_{\ell+1}^{(k)},\cdots, T_m^{(k)}\right)\right].
\end{eqnarray*}
The condition (\ref{eq2}) is also satisfied with $\xi_\ell=1/m$.
$\square$

\medskip


\section{Generalization Error Analyses}
\label{Bounds}

Based on the previous proposition, we can show that at each iteration $k\in [K]$ of the {\CC} algorithm, the Empirical Risk Minimization principle which outputs the classifier $f_k\in\mathcal F_k$ is consistent. This result is stated in Theorem \ref{consistencyCC} which provides bounds on the generalization error of $f_k\in\mathcal F_k, \forall k\in [K]$. The notion of function class capacity used in the bounds, is the Rademacher complexity of the class of functions $\mathcal F_k$ defined as
\begin{equation}
    \label{eq:Rademacher}
\mathfrak{R}_{m}({\cal F}_k)=\BBe_{\Trn_k}\BBe_{\boldsymbol{\sigma}}\left[\sup_{f_k\in {{\cal F}_k}}\frac{2}{m}\left|\sum_{i=1}^m\sigma_i  f_k\left(\z_i^{(k)}\right)\right| | \z^{(k)}_1,\ldots,\z^{(k)}_m \right],
\end{equation}
where $\boldsymbol{\sigma}=(\sigma_i)_{i\in [m]}$, called Rademacher variables are independent uniform random variables taking values in $\{-1,+1\}$; $\forall i\in [m], \BBp(\sigma_i=-1)=\BBp(\sigma_i=+1)=1/2$.

The proof of the theorem is based on the following lemma which bounds $\BBe_{\Trn_k}\left[\sup_{f_k \in {\cal F}_k} \left( {{\cal L}}(f_k)-\hat{\cal L}_m(f_k)\right)\right]$ with respect to the Rademacher complexity of the class of functions $\mathcal{F}_k$, described in Equation \eqref{eq:Rademacher}, and the dependency measure $\gamma^{(k)}$ of Definition \ref{defalphabeta}.

\begin{pro}\label{lem2}
For any random pair of example $(X,\boldsymbol{Y}=(Y^{(k)})_{k\in[K]})\in\Inp\times\Outp$ drawn i.i.d with respect to a probability distribution $\Dist$, $(Z^{(k)},Y^{(k)})\in \bar{\X}_k\times \{-1,+1\}$ the associated pair; with $Z^{(k)}$ the vector formed by $X$ augmented by predictions provided by the $k-1$ previous classifiers of the \CC{} algorithm. Let us denote $Z^{'(k)}$ an independent copy of $Z^{(k)}$. Under the previous notations, we have
\begin{eqnarray*}
&&\BBe_{\Trn_k}\left[\sup_{f_k \in {\cal F}_k} \left({{\cal L}}(f_k)-\hat{\cal L}_m(f_k) \right)\right]\leq  \mathfrak{R}_{m}(\ell_{0/1}\circ{\cal F}_k) + \Pi_{k}+ \gamma_1^{(k)},
\end{eqnarray*}
where $\ell_{0/1}: (y,\hat y)\mapsto \mathbbm{1}_{y\neq \hat y}$ is the $0-1$ loss, and 
\[
\Pi_k=\sup_{f_k\in {\cal F}_k}\frac{1}{m}\!\sum_{i=1}^m\left|\BBe_{Z_i^{'(k)}}\BBe_{Y_i^{(k)}}\!\left[\ell_{0/1}\left(f_k(Z_i^{'(k)}), Y_i^{(k)}\right)\right]\!-\! \BBe_{(Z_i^{(k)},Y_i^{(k)})}\!\left[\ell_{0/1}\left(f_k(Z_i^{(k)}), Y_i^{(k)}\right)\right]\right|.
\]
\end{pro}

\textbf{Proof. } Let $(X_i,(Y_i^{(s)})_{s\in [k]})_{i\in [m]}\in(\Inp\times\Outp)^m$ be $m$ i.i.d. random variables generated from  the distribution $\Dist$. 
We have,
\begin{equation*}
\BBe_{\Trn_k}\! \left[\sup_{f_k \in {\cal F}_k} \left({{\cal L}}(f_k)-\hat{\cal L}_m(f_k) \right)\right]\!=\!\!
\int \!\BBe_{(Y_i^{(k)})_{i\in[m]}}\left[\Phi\left(T_1^{(k)},..,T_m^{(k)}\right)|\left(\obs_j,\left(y_j^{(s)}\right)_{s\in [k-1]}\right)_{j\in [m]}\right]d\mathbb{P}(\obs_i,(y_i^{(s)})_{s\in[k-1],\, i\in[m]}),
\end{equation*}
where $\mathbb{P}$ is the distribution of $\{(X_i,(Y_i^{(s)})_{1\leq s\leq k-1}),\,\, 1\leq i\leq m\}$.
Now, consider $\left(X'_i,(Y_i^{'(s)})_{s\in [k-1]}\right)_{i\in [m]}$, an independent copy  of $\left(X_i,(Y_i^{(s)})_{s\in [k-1]}\right)_{i\in [m]}$ which is also  independent from $\left(Y_i^{(k)}\right)_{i\in [m]}$.  
Following the steps of the proof of Proposition \ref{lem1} it comes,
\begin{equation*}
\left|\BBe_{(Y_i^{(k)})_{i\in[m]}}\left[\Phi({T}_{1}^{(k)},\cdots, {T}_m^{(k)})|\left(\obs_i,\left(y_i^{(s)}\right)_{1\leq s\leq k-1}\right)_{1\leq i\leq m}\right]- \BBe_{(Y_i^{(k)})_{i\in[m]}}\left[\Phi( T_{1}^{'(k)},\cdots, T_m^{'(k)})\right]\right|\leq \gamma_1^{(k)}.
\end{equation*}
 Consequently, we get
\begin{equation}
\label{pro2-1}
\BBe_{\Trn_k}\left[\sup_{f_k \in {\cal F}_k} \left({{\cal L}}(f_k)-\hat{\cal L}_m(f_k)\right)\right]\leq \gamma_1^{(k)} + \int \BBe_{(Y_i^{(k)})_{i\in[m]}}\left[\Phi(T_{1}^{'(k)},\cdots, T_m^{'(k)})\right]d\mathbb{P}\left(\obs_i,(y_i^{(s)})_{s\in[k-1],\, i\in[m]}\right),
\end{equation}
where $\forall i\in[m], T_i^{'(k)}=\ell_{0/1}\left(f_k\left(\z_i^{'(k)}\right),Y_i^{(k)}\right)$, and each $\z_i^{'(k)}, i\in [m]$ is obtained from $\left(\obs'_i,(y_i^{'(s)})_{s\in [k-1]}\right)_{i\in [m]}$ and the predictions of the $k-1$ previous classifiers resulted by the {\CC} algorithm.  Now if we look at the second term of equation (\ref{pro2-1}), we have
\begin{align}
& \int \BBe_{(Y_i^{(k)})_{i\in[m]}}\left[\Phi(T_{1}^{'(k)},\cdots, T_m^{'(k)})\right]d\mathbb{P}\left(\obs_i,(y_i^{(s)})_{s\in[k-1],\, i\in[m]}\right) \label{eq:ExpPhi}\\
=&\!\!\int \!\!\BBe_{(Y_i^{(k)})_{i\in[m]}}\left[\sup_{f_k\in {\cal F}_k}\frac{1}{m}\sum_{i=1}^m\left(\BBe_{(Z_i^{(k)},Y_i^{(k)})}\left[\ell_{0/1}\left(f_k(Z_i^{(k)}), Y_i^{(k)}\right)\right] - \ell_{0/1}\left(f_k(\z_i^{'(k)}), Y_i^{(k)}\right)\right)\right] d\BBp\left((\obs_i,y_i^{(s)})_{s\in [k-1],\,j\in [m]}\right).\nonumber
\end{align}
By adding and subtracting $\BBe_{Y_i^{(k)}}\left[\ell_{0/1}\left(f_k(\z_i^{'(k)}), Y_i^{(k)}\right)\right]$ in the second term of equation (\ref{eq:ExpPhi}), and using the triangle inequality for the $\ell_\infty$ norm it comes,
\begin{align}\label{eq:FInt}
&\displaystyle \int \BBe_{(Y_i^{(k)})_{i\in[m]}}\left[\Phi(T_{1}^{'(k)},\cdots, T_m^{'(k)})\right]d\mathbb{P}\left(\obs_i,(y_i^{(s)})_{s\in[k-1],\, i\in[m]}\right)\\
 \leq &\displaystyle \int \BBe_{(Y_i^{(k)})_{i\in[m]}}\left(\sup_{f_k\in {\cal F}_k}\frac{1}{m}\left|\sum_{i=1}^m\left(\ell_{0/1}\left(f_k(\z_i^{'(k)}), Y_i^{(k)}\right)-\BBe_{Y_i^{(k)}}\left[\ell_{0/1}\left(f_k(\z_i^{'(k)}), Y_i^{(k)}\right)\right]\right)\right|\right) d\mathbb{P}\left(\obs_i,(y_i^{(s)})_{s\in[k-1],\, i\in[m]}\right)\nonumber\\
+ &\displaystyle\!\! \int \!\!\sup_{f_k\in {\cal F}_k}\frac{1}{m}\sum_{i=1}^m\left|\BBe_{Y_i^{(k)}}\left[\ell_{0/1}\left(f_k(\z_i^{'(k)}), Y_i^{(k)}\right)\right]- \BBe_{(Z^{(k)}_i,Y_i^{(k)})}\left[\ell_{0/1}\left(f_k(Z_i^{(k)}), Y_i^{(k)}\right)\right]\right|d\mathbb{P}\left(\obs_i,(y_i^{(s)})_{s\in[k-1],\, i\in[m]}\right).\nonumber
\end{align}

Recall that $(X'_i,(Y_i^{'(s)})_{s\in [k-1]})_{i\in [m]}$ is an independent copy of $(X_i,(Y_i^{(s)})_{s\in [k-1]})_{i\in [m]}$ and is independent of  $(Y_i^{(k)})_{i\in [m]}$. Hence, $\left(\ell_{0/1}\left(f_k(\z_i^{'(k)}), Y_i^{(k)}\right)-\BBe_{Y_i^{(k)}}\left[\ell_{0/1}\left(f_k(\z_i^{'(k)}), Y_i^{(k)}\right)\right]\right)_{i\in [m]}$
are also independent and centered. 
The symmetrization lemma \cite[p.12]{PenaG99}, \cite[p. 36]{Mohri:2012}, applied to those random variables bounds in the first term of the above inequality gives,
\begin{multline}\label{eq:Sym}
\BBe_{(Y_i^{(k)})_{i\in[m]}}\left(\sup_{f_k\in {\cal F}_k}\frac{1}{m}\left|\sum_{i=1}^m\left(\ell_{0/1}\left(f_k(\z_i^{'(k)}), Y_i^{(k)}\right)- \BBe_{Y_i^{(k)}}\left[\ell_{0/1}\left(f_k(\z_i^{'(k)}), Y_i^{(k)}\right)\right]\right)\right|\right)\\ 
 \leq 2\BBe_{ (Y_i^{(k)})_{i\in[m]}} \BBe_{\boldsymbol{\sigma}}\left(\sup_{f_k\in {\cal F}_k}\frac{1}{m}\left|\sum_{i=1}^m \sigma_i\ell_{0/1}\left(f_k(\z_i^{'(k)}), Y_i^{(k)}\right)\right|\right)
\end{multline}
The result of the lemma follows from equations \eqref{pro2-1}, \eqref{eq:FInt}, \eqref{eq:Sym}, and by integrating with respect to the distribution of $(X_i,(Y^{(s)}_i)_{s\in[k-1]})_{i\in [m]}$. $\Box$

From Propositions \ref{lem1} and \ref{lem2}, we can derive a Rademacher generalization bound for any prediction function found at each step $k\in[K]$ of the {\CC} algorithm  as stated below.
\begin{theo}
\label{consistencyCC}
Based on the notations of Proposition \ref{lem1}; for any $1>\delta>0$ and any $f_k\in\mathcal{F}_k$ found by the {\CC} algorithm at step $k$, the following generalization bound holds with probability at least $1-\delta$~:
\begin{equation}
    \label{eq:GenBound}
    \mathcal{L}(f_k)\leq \hat{\mathcal{L}}_m(f_k)+ \gamma_1^{(k)}+\rho_k+\mathfrak{R}_{m}(\ell_{0/1}\circ {\cal F}_k)+\sqrt{\frac{s^{(k)}\log \frac{1}{\delta}}{2m^2} },
\end{equation}
where $s^{(k)}=\displaystyle \sum_{\ell=1}^m\left(1+2m\gamma_{\ell+1}^{(k)}\right)^2$.
\end{theo}

\textbf{Proof. } Resolving $\delta=\exp{-\frac{2m^2\epsilon^2}{\sum_{\ell=1}^m \left(1+2m\gamma^{(k)}_{\ell}\right)^2}}$ for $\epsilon$, we have from Proposition \ref{lem1} that with probability at least $1-\delta$,
\begin{equation*}
    \forall f_k\in\mathcal{F}_k, \mathcal{L}(f_k)-\hat{\mathcal{L}}_m(f_k)\leq \sup_{f_k \in\mathcal F_k} \left({{\cal L}}(f_k)-\hat{\cal L}_m(f_k)\right)\leq \BBe_{\Trn_k}\left[\sup_{f_k \in \mathcal F_k} \left({{\cal L}}(f_k)\right)-\hat{\cal L}_m(f_k) \right] + \sqrt{\frac{s^{(k)}\log \frac{1}{\delta}}{2m^2} }.
\end{equation*}
From Proposition \ref{lem2} it then comes that with probability at least $1-\delta$,
\begin{equation}
    \label{eq:Borne2}
    \forall f_k\in\mathcal{F}_k, \mathcal{L}(f_k)-\hat{\mathcal{L}}_m(f_k)\leq   \mathfrak{R}_{m}(\ell_{0/1}\circ{\cal F}_k) + \Pi_{k}+ \gamma_1^{(k)} + \sqrt{\frac{s^{(k)}\log \frac{1}{\delta}}{2m^2} }, 
\end{equation}
where 
\[
\Pi_k=\sup_{f_k\in {\cal F}_k}\frac{1}{m}\!\sum_{i=1}^m\left|\BBe_{Z_i^{'(k)}}\BBe_{Y_i^{(k)}}\!\left[\ell_{0/1}\left(f_k(Z_i^{'(k)}), Y_i^{(k)}\right)\right]\!-\! \BBe_{(Z_i^{(k)},Y_i^{(k)})}\!\left[\ell_{0/1}\left(f_k(Z_i^{(k)}), Y_i^{(k)}\right)\right]\right|.
\]
The purpose now is to control $\Pi_k$, by bounding the term $\left|\BBe_{Z_i^{'(k)}}\BBe_{Y_i^{(k)}}\!\left[\ell_{0/1}\left(f_k(Z_i^{'(k)}), Y_i^{(k)}\right)\right]\!-\! \BBe_{(Z_i^{(k)}, Y_i^{(k)})}\!\left[\ell_{0/1}\left(f_k(Z_i^{(k)}), Y_i^{(k)}\right)\right]\right|$, we observe that 
\begin{multline}\label{pp}
\BBe_{Z_i^{'(k)}}\BBe_{Y_i^{(k)}}\!\left[\ell_{0/1}\left(f_k(Z_i^{'(k)}), Y_i^{(k)}\right)\right] - \BBe_{(Z_i^{(k)}, Y_i^{(k)})}\!\left[\ell_{0/1}\left(f_k(Z_i^{(k)}), Y_i^{(k)}\right)\right] = \nonumber\\
\sum_{y_i^{(k)}} \left(\BBe_{Z_i^{'(k)}}\BBe_{Y_i^{(k)}}\left[\mathbbm{1}_{Y_i^{(k)}= y_i^{(k)}}\ell_{0/1}(f_k(Z_i^{'(k)}), y_i^{(k)})\right]-\BBe_{( Z_i^{(k)},Y_i^{(k)})}\left[\mathbbm{1}_{Y_i^{(k)}= y_i^{(k)}}\ell_{0/1}\left(f_k(Z_i^{(k)}), y_i^{(k)}\right)\right]  \right){\nonumber}
\end{multline}
We have, by the independence of $Y_i^{(k)}$ and $f_k(Z_i^{'(k)})$, 
\begin{multline*}
\sum_{y_i^{(k)}} \left(\BBe_{Z_i^{'(k)}}\BBe_{Y_i^{(k)}}\left[\mathbbm{1}_{Y_i^{(k)}= y_i^{(k)}}\ell_{0/1}(f_k(Z_i^{'(k)}), y_i^{(k)})\right]-\BBe_{( Z_i^{(k)},Y_i^{(k)})}\left[\mathbbm{1}_{Y_i^{(k)}= y_i^{(k)}}\ell_{0/1}\left(f_k(Z_i^{(k)}), y_i^{(k)}\right)\right]  \right){\nonumber} \\
 = \sum_{y_i^{(k)}} \left(\BBe_{Y_i^{(k)}}[\mathbbm{1}_{Y_i^{(k)}= y_i^{(k)}}]\BBe_{Z_i^{'(k)}}\left[\ell_{0/1}\left(f_k(Z_i^{'(k)}, y_i^{(k)})\right)\right]-\BBe_{( Z_i^{(k)},Y_i^{(k)})}\left[\mathbbm{1}_{Y_i^{(k)}= y_i^{(k)}}\ell_{0/1}\left(f_k(Z_i^{(k)}), y_i^{(k)}\right)\right]  \right).
\end{multline*}
and since $f_k(Z_i^{'(k)})$ and $f_k(Z_i^{(k)})$ are identically distributed
\begin{multline*}
\sum_{y_i^{(k)}} \left(\BBe_{Y_i^{(k)}}[\mathbbm{1}_{Y_i^{(k)}= y_i^{(k)}}]\BBe_{Z_i^{'(k)}}\left[\ell_{0/1}\left(f_k(Z_i^{'(k)}, y_i^{(k)})\right)\right]-\BBe_{( Z_i^{(k)},Y_i^{(k)})}\left[\mathbbm{1}_{Y_i^{(k)}= y_i^{(k)}}\ell_{0/1}\left(f_k(Z_i^{(k)}), y_i^{(k)}\right)\right]  \right){\nonumber} \\
 = \sum_{y_i^{(k)}} \left(\BBe_{Y_i^{(k)}}[\mathbbm{1}_{Y_i^{(k)}= y_i^{(k)}}]\BBe_{Z_i^{(k)}}\left[\ell_{0/1}\left(f_k(Z_i^{(k)}, y_i^{(k)})\right)\right]-\BBe_{( Z_i^{(k)},Y_i^{(k)})}\left[\mathbbm{1}_{Y_i^{(k)}= y_i^{(k)}}\ell_{0/1}\left(f_k(Z_i^{(k)}), y_i^{(k)}\right)\right]  \right).
\end{multline*}
Now since  $f_k(Z_i^{(k)})$ is $(X_i,(Y_i^{(s)})_{s\in [k-1]})_{i\in [m]}$-measurable,
\begin{equation*}
\BBe_{( Z_i^{(k)},Y_i^{(k)})}\left[\mathbbm{1}_{Y_i^{(k)}= y_i^{(k)}}\ell_{0/1}\left(f_k(Z_i^{(k)}), y_i^{(k)}\right)\right]= \BBe_{Z_i^{(k)}}\left[\ell_{0/1}\left(f_k(Z_i^{(k)}), y_i^{(k)}\right)\BBe_{Y_i^{(k)}}\left(\mathbbm{1}_{Y_i^{(k)}= y_i^{(k)}}\left|\right. (X_i,(Y_i^{(s)})_{s\in [k-1]})_{i\in [m]}\right)\right],
\end{equation*}
From the last two inequalities, it comes
\begin{multline*}\label{pp}
\BBe_{Z_i^{'(k)}}\BBe_{Y_i^{(k)}}\!\left[\ell_{0/1}\left(f_k(Z_i^{'(k)}), Y_i^{(k)}\right)\right] - \BBe_{(Z_i^{(k)}, Y_i^{(k)})}\!\left[\ell_{0/1}\left(f_k(Z_i^{(k)}), Y_i^{(k)}\right)\right] =\\
\sum_{y_i^{(k)}} \BBe_{Z_i^{(k)}}\left[\ell_{0/1}\left(f_k(Z_i^{(k)}), y_i^{(k)}\right)\left(\BBe_{Y_i^{(k)}}[\mathbbm{1}_{Y_i^{(k)}= y_i^{(k)}}]-\BBe_{Y_i^{(k)}}\left(\mathbbm{1}_{Y_i^{(k)}= y_i^{(k)}}\left|\right. (X_i,(Y_i^{(s)})_{s\in [k-1]})_{i\in [m]}\right)\right)\right].
\end{multline*}
Since $\forall Z_i^{(k)}, \ell_{0/1}\left(f_k(Z_i^{(k)}), y_i^{(k)}\right)\leq 1$, it finally comes 
\begin{equation}
\Pi_k \leq \sup_{1\leq i\leq m} \sum_{y_i^{(k)}} \left|\BBe_{Y_i^{(k)}}[\mathbbm{1}_{Y_i^{(k)}= y_i^{(k)}}]-\BBe_{Y_i^{(k)}}\left(\mathbbm{1}_{Y_i^{(k)}= y_i^{(k)}}\left|\right. (X_i,(Y_i^{(s)})_{s\in [k-1]})_{i\in [m]}\right)\right|\leq \rho_k.
\end{equation}
The last bound together with Eq. (\ref{eq:Borne2}) completes the proof of Theorem \ref{consistencyCC}. $\Box$

\textbf{Comments.} We now give explicit bounds in two important cases : the independent and the Markovian framework. 

We first assume that we have only one label and that the training examples are identically and independently distributed. Then, we have $\rho_k=\gamma_1^{(k)}=0, \text{ and then } s^{(k)}=m, \forall k\in[K]$ and recover the classical Rademacher bounds. Our result is then an extension of generalization bounds in the independent setting. Here we added a novel term related to dependence between the successive classifiers of the chain.

In the Markovian case, the coefficients $\rho^{(k)}$ and $\gamma_1^{(k)}$ involved in the bound are equal. Since they can be estimated empirically from the data, our result paves the way to practical applications, allowing for example to compare chaining strategies.

\section{Conclusion and perspectives}
\label{Conclusion}
In this paper we proposed Rademacher generalization bounds for classifiers found at each iteration of the {\CC} algorithm. In this framework, the learning is carried out by first defining an order on the labels and then learning classifiers sequentially, following the chain. At each iteration, the learner receives a labeled sample based on the initial training set, augmented by the previous classifiers predictions, and trains a classifier following the \texttt{ERM} principle. 


Many empirical studies have shown the benefits of this approach for multi-label classification. In particular, different strategies have been proposed to define the best chaining of classifiers, or to combine efficiently different classifier chains. However, to the best of our knowledge there is no theoretical study related to the generalization ability of classifiers produced by the {\CC} algorithm. The main challenge is that, at each iteration of the algorithm, a new classifier is found using all the training samples, making the augmented feature representation of observations to be interdependent. In this case, usual techniques developed to derive generalization error bounds under the i.i.d. assumption cannot be deployed to measure the deviation between the generalization error and the empirical risk of the learned classifiers.      

The contribution of this paper is a new framework, that provides a new insight on the generalization ability of classifiers produced by the {\CC} algorithm. Our analysis is based on large deviation inequalities proposed by \cite{RIO2000905} that extends McDiarmid inequality \cite{McDiarmid:89} to the case of dependently distributed random variables. It involves two dependency coefficients, where the first one evaluates, the variational distance between the probability measure of the current class in the chain conditionally to the whole training examples and their previous class labels in the chain, and the distribution of the current class. The second coefficient estimates the variational distance between the same probability measures over any subset of the training samples of different sizes. Based on these coefficients, we derive a generalization bound which involves a Rademacher type of complexity term defined over interdependent training samples where the vector representations of observations are augmented at each iteration of the {\CC} algorithm. The proposed bound extends Rademacher complexity bounds for the binary classification where the training examples are identically and independently distributed.

In the Markovian case, these coefficients can be estimated from the data. Future works will be devoted to a comparison of chaining strategies and design of new ones, to decide the order of the chain.

\newpage


\end{document}